\documentclass[11pt]{article}

\usepackage[preprint]{acl}
\usepackage{times}
\usepackage{latexsym}
\usepackage[T1]{fontenc}
\usepackage[utf8]{inputenc}
\usepackage{microtype}
\usepackage{amsmath,amssymb}
\usepackage{graphicx}
\usepackage{booktabs}
\usepackage{xcolor}

\title{When Does Streaming Tool Use Help?\\
Characterizing Tool-Intent Stabilization in Streaming Retrieval-Augmented Generation}
\author{Elroy Galbraith \\
  SMG Labs \\
  \texttt{elroy.galbraith@gmail.com}}

\newcommand{\nTotal}{2706}
\newcommand{\nSplitZero}{1371}
\newcommand{\nGroundPct}{35.4\%}      % groundable %
\newcommand{\nRetrPct}{21.3\%}        % gold retrieved in top-k at some prefix (k=3) %
\newcommand{\nFallbackPct}{78.7\%}    % fallback (t_sc) population = 100% - nRetrPct %
\newcommand{\phiScMean}{0.67}
\newcommand{\phiScMed}{0.73}
\newcommand{\phiSufMean}{0.26}
\newcommand{\phiSufMed}{0.14}
\newcommand{\volMean}{0.52}
\newcommand{\volSharePct}{20.9\%}
\newcommand{\streamPct}{73.9\%}
\newcommand{\rqThreeN}{60}
\newcommand{\measSaved}{1122}        % L=600 measured mean %
\newcommand{\hPred}{590}             % L=600 H mean %
\newcommand{\rhoLsix}{+0.14}         % Spearman(H, measured) at L=600 %
\newcommand{\measSavedLk}{1457}      % L=1000 measured mean %
\newcommand{\hPredLk}{950}           % L=1000 H mean %
\newcommand{\rhoLk}{+0.12}           % Spearman(H, measured) at L=1000 %
\newcommand{\nSatLsix}{59}           % L=600: queries pinned at H=600 (of 60) %
\newcommand{\nTailLk}{7}             % L=1000: queries below the H cap (of 60) %
\newcommand{\streamSufPct}{95.2\%}   % streamable among retrieved-gold, using t_suf (n=292) %
\newcommand{\streamScPct}{68.1\%}    % streamable among fallback, using t_sc (n=1079) %
\newcommand{\kwAllH}{17.0}           % Kruskal-Wallis H, all 8 types %
\newcommand{\kwAllP}{0.017}
\newcommand{\kwBigH}{13.9}           % Kruskal-Wallis H, 5 classes with n>=10 %
\newcommand{\kwBigP}{0.008}
\newcommand{\epsSq}{0.06}            % KW rank effect size eps^2 = H/(n-1), all 8 types; stats.py %
\newcommand{\epsSqBig}{0.05}         % KW eps^2 = H/(n-1), 5 classes with n>=10; stats.py %
\newcommand{\qgenMs}{590}            % query generation %
\newcommand{\fuseMs}{2520}           % fuse/reflect %
\newcommand{\groundFuzzyPct}{43.3\%} % groundable under fuzzy (vs 35.4 exact) %
\newcommand{\retrFuzzyPct}{24.4\%}   % retrieved-gold under fuzzy (vs 21.3 exact) %
\newcommand{\phiSufFuzzy}{0.281}     % phi_suf mean under fuzzy (vs 0.264 exact) %
\newcommand{\nSetFuzzy}{20}          % 'set' support under fuzzy (vs 2 exact) %
\newcommand{\negRateSuf}{1.7\%}      % net-negative-saving rate, retrieved-gold pop %
\newcommand{\negRateSc}{1.7\%}       % net-negative-saving rate, t_sc fallback pop %
\newcommand{\phiScReplay}{0.68}      % mean phi of the t_sc replay sample %
\newcommand{\retrDensePct}{26.5\%}   % retrieved-gold, dense (vs 21.3 BM25) %
\newcommand{\phiScDense}{0.85}       % phi_sc mean, dense (vs 0.67) %
\newcommand{\phiSufDense}{0.31}      % phi_suf mean, dense (vs 0.26) %
\newcommand{\phiSufDenseMed}{0.22}   % phi_suf median, dense (vs 0.14) %
\newcommand{\rhoDenseType}{0.79}     % Spearman(BM25,dense) over phi_suf type ranking %
\begin{document}
\maketitle

\begin{abstract}
Streaming Retrieval-Augmented Generation (Streaming RAG) hides tool latency by
issuing retrieval queries in parallel with the user's still-arriving input, before
the utterance is complete. Speculation can only help, though, when the correct
query becomes determinable before the user stops speaking or typing---a property of
the query, not the system. We name and measure this property,
\textbf{tool-intent stabilization}: the point in the input stream at which a
speculative query's retrieval converges on the answer-bearing result. On the CRAG
benchmark (\nSplitZero{} validation questions) we (i) characterize how
stabilization is distributed across queries; (ii) derive a model-agnostic bound $H$
on the share of tool latency hideable behind the remaining input, given tool
latency $L$ and input cadence $\delta$; (iii) validate it against a working
streaming pipeline; and (iv) ask which query properties predict early versus late
stabilization. Stabilization is typically early: at a realistic operating point a
\streamPct{} streamable fraction of the benchmark admits latency hiding, and $H$
acts as a conservative aggregate floor that realized savings meet or exceed---though
it does not predict savings query by query. Question type yields a statistically
significant but small early/late split. The study needs no model training and runs
on commodity CPU hardware; a dense-retriever replication confirms the
early-stabilization effect is not a BM25 lexical artifact.
\end{abstract}

\section{Introduction}
Tool-augmented dialogue systems increasingly rely on external retrieval to remain
factual, but tool calls add latency that disrupts conversational flow. Streaming
RAG~\cite{arora2025streamrag} addresses this by predicting and issuing tool
queries in parallel with the user's still-arriving input, then reflecting
on whether retrieved results suffice. The approach reports sizable accuracy and
latency improvements and is modality-agnostic, applying equally to typed input.

The reported latency benefit, however, is an aggregate over a benchmark. It
leaves a basic question unanswered: \textbf{for which queries can speculation
help at all, and by how much?} The answer is not a property of the model, the
retriever, or the speech stack. It is a property of where, within an utterance,
the information that determines the tool query first appears. If the
decisive term arrives last (``who makes the console called the
\emph{PlayStation}''), no early speculative query can retrieve the right evidence,
and the latency win collapses to zero regardless of system quality. If the
decisive term arrives early, almost the entire tool latency can be hidden behind
the remaining input. We initially expected these two regimes to show up as a
\emph{bimodal} distribution of stabilization; empirically (\S5.1) the picture is
better described as a large early-stabilizing mass with a thin late tail---the
early case is common and the late case is the rarer failure mode---and we report
the distribution as found rather than as hypothesized.

We study this directly. By characterizing tool-intent stabilization as a
measurable, query-intrinsic quantity, we obtain (i) a model-agnostic
\emph{aggregate} bound on the share of tool latency that can be hidden behind
ongoing input, computable \emph{before} any system is built; (ii) a principled
account of the failure mode; (iii) a coarse build/skip signal---weak at the level
of an individual query---for whether a learned trigger is worth its cost; and
(iv) results that transfer across text and speech because the quantity is
modality-agnostic. Our contribution is measurement and analysis rather than a new
system.

\section{Background and Related Work}
Retrieval-Augmented Generation~\cite{lewis2020rag} grounds generation in
retrieved evidence; dense passage retrieval~\cite{karpukhin2020dpr} and the
sparse BM25 model~\cite{robertson2009bm25} are the standard retrieval backbones.
Streaming RAG~\cite{arora2025streamrag} adds a speculative dimension: a
\emph{Trigger} decides when to fire a tool query from a partial utterance,
multiple speculative \emph{Threads} run in parallel, and a \emph{Reflector}
judges whether results suffice. The idea is closely related in spirit to
speculative decoding~\cite{leviathan2023speculative}, which hides latency by
doing work that may be discarded; here the discarded work is a premature tool
call. Our analysis is orthogonal to the system: we quantify the headroom that any
such system can exploit, set by the query alone. Two literatures tell us why such
headroom should exist and how to measure when a prefix is safe to act on.

\subsection{Incremental comprehension and anticipation}
Why expect tool intent to settle before a query is complete? Human comprehension
already works this way. That the referential
intent of an utterance is fixed incrementally---often well before the utterance
ends---is among the most robust findings in psycholinguistics.
\citet{altmann1999incremental} showed that a verb such as ``eat'' directs
anticipatory attention to edible objects before they are named, so partial input
already restricts what can follow. \citet{kamide2003timecourse} extend this
cross-linguistically: in head-final Japanese, case-marked pre-verbal arguments let
comprehenders anticipate a forthcoming argument even before the verb (the head) is
heard---\emph{pre-head} anticipation~\cite{kamide1999prehead}. This is the human
analogue of firing a tool query from a prefix: \emph{where} the disambiguating
cue sits---varying with word order and case morphology---governs how early intent
stabilizes.

\subsection{Stabilization over a growing prefix}
If comprehension fixes intent early, an engineered system faces the dual problem:
deciding when a prediction over a growing input prefix becomes safe to commit. This
is the central question of \emph{simultaneous} and \emph{incremental} NLP. Simultaneous machine
translation must decide, token by token, whether enough source has arrived to emit
output: wait-$k$ and prefix-to-prefix policies~\cite{ma2019stacl} and learned
read/write agents~\cite{gu2017simultaneous} trade latency against the risk of
committing before the disambiguating word arrives---precisely our $L$-versus-residual
tension. \citet{grissom2014dont} cast the same problem as an agent that waits until
it can confidently predict the sentence-final verb before committing; its failure
mode, premature commitment forcing revision, is exactly what our volatility $V$
records. Closest to our measurement is incremental spoken language understanding:
\citet{devault2011incremental} build a meaning frame from partial ASR and quantify
understanding as a function of prefix length, and the incremental-unit
framework~\cite{schlangen2011incremental} formalizes the same prefix-monotonicity
and revision concerns. These efforts target a \emph{closed} intent or dialogue-act
ontology. Our contribution is to port the lens to \emph{tool/retrieval} intent over
an open-world knowledge benchmark: rather than learning a commit policy, we
\emph{measure} the query-intrinsic point at which the retrieval target stabilizes,
which upper-bounds what any such policy could hide.

\section{Problem Formalization}
Let a query $Q$ be a word sequence $q_{1:n}$ of length $n$ words, and let $q_{1:t}$
denote the prefix of the first $t$ words (we measure stabilization in words
throughout, matching the whitespace tokenization used by both the retriever and
the input-cadence model below). Let $R$ be a retriever returning a
ranked document list; write $d_t = \mathrm{top\text{-}1}(R(q_{1:t}))$ for the top
document retrieved from the prefix, $d_n$ for the full-query top document, and
$d^\star$ for the gold answer-bearing document.

\paragraph{Self-consistency stabilization $t_{\mathrm{sc}}$.} the smallest $t$
such that for all $s \in \{t, \ldots, n\}$, $d_s = d_n$: every prefix from $t$
onward retrieves the same top document as the full query.

\paragraph{Sufficiency stabilization $t_{\mathrm{suf}}$.} the smallest $t$ such
that $d^\star \in \mathrm{top\text{-}}k(R(q_{1:t}))$: the prefix already surfaces
the gold evidence. This is the operationally meaningful notion, since the full
query itself may retrieve imperfectly.

\paragraph{Stabilization fraction $\phi = t^\star/n \in (0,1]$,} reported for both
$t_{\mathrm{sc}}$ and $t_{\mathrm{suf}}$. Lower $\phi$ means earlier stabilization
and more headroom to hide latency.

\paragraph{Stabilization volatility $V$,} the number of top-1 changes occurring
\emph{after} $d_n$ is first reached. High $V$ signals early-commit risk: a
confident-looking early result that a later token would overturn.

\paragraph{Hidden latency.}
\begin{equation}
H(Q; L, \delta) = \min\!\Big(L,\; \max\!\big(0,\; \tfrac{n - t^\star}{\delta}\big)\Big),
\end{equation}
where $L$ is tool latency (ms) and $\delta$ is the input cadence in words per
second, so that $(n-t^\star)/\delta$ is the residual input time (in seconds, scaled
to ms) still to arrive after the query has stabilized. $H$ is the portion of tool
latency that completes behind the user's remaining input. It bounds the
\emph{input-hideable} component of the per-query perceived-latency saving---not the
total saving, since a real pipeline can additionally overlap other costs (\S5.4).
A slower cadence (smaller $\delta$) lengthens the residual and so raises $H$ when
tool latency is the larger term.

\paragraph{Streamable.} $\mathrm{Streamable}(Q; L, \delta, \theta)$ holds iff
$H \geq \theta L$ for a coverage threshold $\theta$ (e.g.\ $\theta=0.8$ hides at
least 80\% of tool latency). The streamable fraction of a workload is the central
deployment-relevant statistic.

\paragraph{Research questions.} We address four pre-registered questions.
\textbf{RQ1} asks how $\phi$ is distributed across queries; we expect a bimodal
shape with an early-stabilizing mass and a late-stabilizing tail.
\textbf{RQ2} asks what fraction of queries admit latency hiding as $L$, $\delta$,
and $\theta$ vary; we expect that for large $L$ the binding constraint is the
residual input time $(n-t^\star)/\delta$, so a \emph{slower} cadence
paradoxically \emph{raises} the streamable fraction.
\textbf{RQ3} asks whether the $H$ bound matches a working pipeline.
\textbf{RQ4} asks which query features predict early vs.\ late stabilization; we
expect question type to be the dominant signal, with simple lookups stabilizing
earlier than multi-hop or aggregation queries.
We address RQ1--RQ4 in \S5 and include top-$k$ and dense-retriever robustness
arms (Table~\ref{tab:topk}, \S\ref{sec:robust}) as sensitivity checks. A
cross-linguistic condition is left to future work. We report
disconfirmations alongside confirmations: RQ2 is confirmed, but the RQ1
distribution is right-skewed rather than bimodal (\S5.1), and the RQ4 ordering
is governed by entity position rather than reasoning complexity (\S5.2).

\section{Method}\label{sec:method}
\paragraph{Data.} We use CRAG, the Comprehensive RAG
Benchmark~\cite{yang2024crag}, Task 1\&2 dev set: \nTotal{} questions, of which
\nSplitZero{} are the validation split we analyze. Each question carries up to
five retrieved web pages, a gold answer (plus alternates), and a question-type
label.

\paragraph{Grounding $d^\star$.} CRAG provides no gold-\emph{passage} label, only
gold answer strings. We derive $d^\star$ by grounding the normalized gold
answer(s) into passage text: substring match for multi-token answers, and
word-boundary match for single-token answers, over the answer and its alternates.
Questions whose answer is never a literal span (false-premise, ``I don't know'',
and many aggregation/dynamic answers) are \emph{ungroundable} and excluded from
$t_{\mathrm{suf}}$; we report the groundable rate explicitly. Because
$t_{\mathrm{sc}}$ requires no grounding, we report it as a grounding-free
robustness check alongside $t_{\mathrm{suf}}$.

String grounding is a recall-oriented heuristic and can produce false positives.
A single-token gold answer (a year, a number, ``yes''/``no'') may match a passage
coincidentally, registering sufficiency before the genuinely answer-bearing
evidence arrives. This biases $t_{\mathrm{suf}}$, and thus $\phi_{\mathrm{suf}}$,
early. Multi-token answers, which require a contiguous
substring match, are far less prone to this. We therefore read $\phi_{\mathrm{suf}}$
as a lower bound on stabilization time on the groundable subset, and treat the
early-stabilization finding as directionally robust rather than exact.

To bound the population-selection bias this strictness induces, we also run a
\emph{relaxed} grounding arm (\S\ref{sec:robust}): for multi-token answers it accepts
a passage if all of the answer's content tokens (stopwords dropped) appear anywhere in
it---bag-of-content-token containment, dropping only the contiguity requirement. This
is deliberately higher-recall and lower-precision than the substring rule, so the two
arms \emph{bracket} the true groundable population and the $\phi_{\mathrm{suf}}$ that
goes with it.

\paragraph{Passages and retrieval.} Page HTML is cleaned to text and chunked into
120-word windows with a 20-word overlap (stride 100), capped at 400 chunks per
question. We index each question's passages with a pure-Python lexical
BM25~\cite{robertson2009bm25} ($k_1{=}1.5$, $b{=}0.75$) and retrieve over every
prefix $q_{1:t}$, $t=1\dots n$, recording $d_t$ and the top-$k$ set at each prefix.
We sweep $k\in\{1,3,5\}$ for the sufficiency definition. All retrieval is
deterministic; the analysis uses the CRAG \texttt{dev\_v4} snapshot (the released
web pages, so retrieval is fixed and not subject to live drift). The two stochastic
or selection steps are seeded and reproducible: the $\rqThreeN{}$-question RQ3
subset is simply the first \rqThreeN{} retrieved-gold questions in dataset order
(no sampling), and the bootstrap confidence intervals (\S5.2) use $10{,}000$
resamples at seed $0$.

\paragraph{Latency model and harness (RQ3).} We stream each query as a uniform
word sequence---a controlled proxy for input cadence $\delta$---and sweep
$(L,\delta,\theta)$ analytically over the grid
$L\in\{100,300,600,1000\}$\,ms, $\delta\in\{2,3,4\}$\,w/s, $\theta\in\{0.5,0.8,1.0\}$
to obtain the streamable surface (RQ2). For RQ3 we replay a subset of questions
through a faithful asynchronous Streaming RAG harness (fixed-interval Trigger,
parallel Threads, Reflector) calibrated with the per-stage latencies reported by
\citet{arora2025streamrag} (Table~3): query generation $\approx\qgenMs{}$\,ms and
fuse/reflect $\approx\fuseMs{}$\,ms, in addition to the swept tool latency $L$. We
compare the \emph{measured} perceived-latency saving (baseline minus streaming)
against the $H$ bound.

\paragraph{Reproducibility.} The pipeline is training-free and CPU-only. Code and
the derived $d^\star$ passage labels (the only non-trivial artifact, since CRAG
ships none) are available at \url{https://github.com/elroy-galbraith/stablize_CRAG}; every per-question metric and the
full $(L,\delta,\theta)$ grid can be recomputed offline without re-running
retrieval.

% ===================================================================== %
\section{Results}
We analyze the \nSplitZero{} validation questions of CRAG Task~1\&2. Under our
string-grounding procedure, \nGroundPct{} of questions are \emph{groundable} (the
gold answer appears verbatim in the retrieved pages); the remainder---dominated by
false-premise items and answers that never occur as a literal span (many
aggregation and dynamic questions)---are excluded from sufficiency statistics. At
$k{=}3$, BM25 surfaces the gold passage in the top-$k$ at some prefix for
\nRetrPct{} of all questions. All reported $\phi_{\mathrm{suf}}$ statistics are
conditional on this retrieved-gold population; $\phi_{\mathrm{sc}}$ is defined for
every question and serves as a grounding-free check.

This retrieved-gold subset is not a random sample of the benchmark: it is the
intersection of questions whose answer is verbatim-groundable \emph{and} whose gold
passage BM25 can surface from some prefix. Both filters plausibly select for
queries with a salient, early lexical anchor, so the conditional
$\phi_{\mathrm{suf}}$ statistics should be read as characterizing this favorable
slice, not ``most answerable questions''; if anything, the selection biases the
reported stabilization \emph{earlier} than it would be on the full workload. We
therefore lean on $\phi_{\mathrm{sc}}$, defined on all \nSplitZero{} questions, as
the grounding-free companion to every sufficiency claim, and we bound the magnitude
of the selection effect with a relaxed-grounding arm below
(\S\ref{sec:robust}): recovering a sixth more of the population moves
$\phi_{\mathrm{suf}}$ by under $0.02$, so the early-stabilization finding survives the
expansion rather than dissolving into it.

\subsection{RQ1: How is stabilization distributed?}
\begin{figure}[t]
\centering
\includegraphics[width=0.78\linewidth]{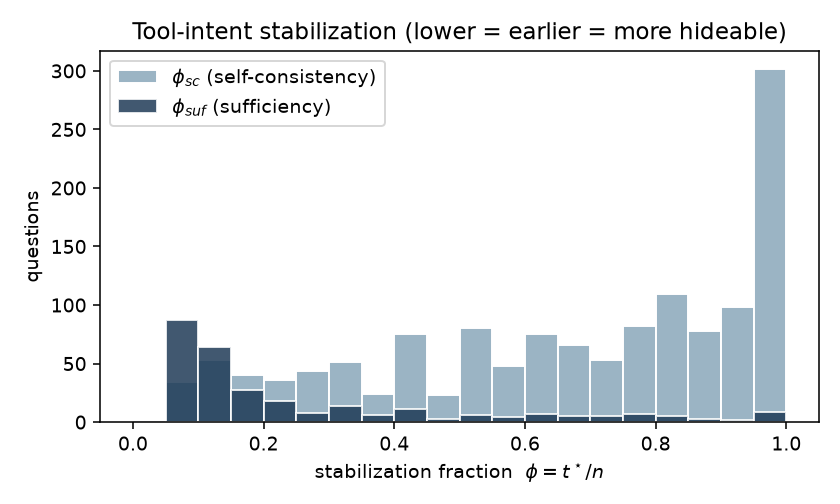}
\caption{Distribution of stabilization fraction $\phi$ at $k{=}3$. Sufficiency
($\phi_{\mathrm{suf}}$, dark) concentrates near zero---gold evidence is retrievable
early---while self-consistency ($\phi_{\mathrm{sc}}$, light) is late and broad.}
\label{fig:phi}
\end{figure}

The two stabilization notions behave very differently (Fig.~\ref{fig:phi}).
Self-consistency stabilizes \emph{late}: the lexical top-1 keeps changing until,
on average, $\phi_{\mathrm{sc}}$ reaches mean \phiScMean{} (median \phiScMed{})---
roughly three-quarters of the way through the query. Sufficiency, in contrast,
stabilizes \emph{early}: the gold passage enters the top-$k$ at mean
$\phi_{\mathrm{suf}}=\phiSufMean{}$ and median \phiSufMed{}, bracketed to
$[\phiSufMean{},\,\phiSufFuzzy{}]$ by a relaxed-grounding arm (\S\ref{sec:robust})---both early. The gap
is the central empirical observation of this study: \textbf{the answer-bearing document is
retrievable from a short prefix long before the query's lexical top-1 settles}.
For streaming tool use this is the favorable regime---one need not wait for the
utterance to ``finish'' for the right evidence to be in hand.

The $\phi_{\mathrm{suf}}$ distribution (Fig.~\ref{fig:phi}) is right-skewed with a
large mass near zero and a thinner late tail, rather than cleanly bimodal as
H1 anticipated. Volatility is modest: the post-stabilization top-1 changes at all
for only \volSharePct{} of questions (mean $V=\volMean{}$), so early-commit risk is
concentrated in a minority of queries rather than pervasive.

\subsection{RQ4: What predicts early vs.\ late stabilization?}
\begin{figure}[t]
\centering
\includegraphics[width=0.92\linewidth]{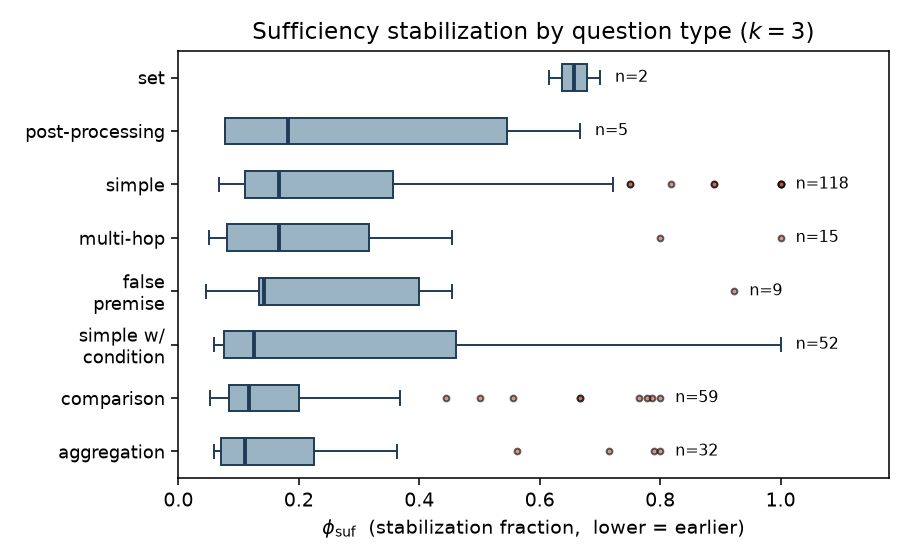}
\caption{Distribution of $\phi_{\mathrm{suf}}$ by question type ($k{=}3$, sorted by
median). The $3\times$ range across medians is real, but box overlaps are large for
adjacent types---only the extremes (\emph{aggregation}/\emph{comparison} earliest,
\emph{set} latest) are reliably ordered. Ordering tracks entity position, not
reasoning complexity.}
\label{fig:bytype}
\end{figure}

Question type produces a coarse early/late split in $\phi_{\mathrm{suf}}$, spanning
a roughly threefold range across medians (Fig.~\ref{fig:bytype}). A Kruskal--Wallis
test across all eight types rejects equality of the $\phi_{\mathrm{suf}}$
distributions ($H{=}\kwAllH{}$, $\mathrm{df}{=}7$, $p{=}\kwAllP{}$); restricting to
the five classes with $n\ge10$---which carry the effect---the test is stronger
($H{=}\kwBigH{}$, $\mathrm{df}{=}4$, $p{=}\kwBigP{}$). The effect, though
significant, is \emph{small}: the rank-based effect size is
$\varepsilon^2{=}\epsSq{}$ across all eight types and $\epsSqBig{}$ for the five
large classes, i.e.\ question type explains on the order of $5$--$6\%$ of the rank
variance in $\phi_{\mathrm{suf}}$. Consistent with a small effect, the interquartile ranges in Fig.~\ref{fig:bytype}
overlap substantially for adjacent types: the ordering is reliable only at the extremes (\emph{aggregation} and
\emph{comparison} earliest, \emph{set} latest---\texttt{set} confirmed at
$n{=}\nSetFuzzy{}$ under the relaxed-grounding arm, \S\ref{sec:robust}), not as a
strict cell-by-cell ranking, and \texttt{post-processing} ($n{=}5$) has too few
points to place reliably. A Dunn's post-hoc with Holm correction
makes this precise: \emph{no} pairwise contrast survives at $\alpha{=}0.05$ (the
closest, \texttt{simple} vs.\ \texttt{aggregation}, reaches $p_{\mathrm{Holm}}{=}0.05$).
The omnibus effect is therefore carried by the overall rank distribution, not by any
single separable pair. We accordingly claim a significant but coarse
type-level effect---a robust early/late \emph{tendency}, not a precise per-type order.
The qualitative ordering refines
hypothesis H3 rather than confirming it. H3 expected simple lookups to stabilize
early and aggregation/comparison/multi-hop to stabilize late. We instead find
\emph{aggregation} and \emph{comparison} among the \emph{earliest}-stabilizing
types, with \emph{set} questions latest. The explanation is that sufficiency
measures when the gold \emph{document} becomes retrievable, not when the answer can
be \emph{computed}: a comparison or aggregation question often names its key
entities up front, so the answer-bearing page is retrievable early even though
producing the final answer requires post-retrieval reasoning. In other words,
retrieval-sufficiency stabilization is governed by entity position in the
utterance, which is not aligned with reasoning complexity. This entity-position
account is carried by the well-populated classes (\texttt{simple},
\texttt{comparison}, \texttt{aggregation}, \texttt{simple\_w\_condition}), so it
does not hinge on the noisy small-$n$ cells. We present it as an explanation
motivated by the data rather than one we have isolated: it yields a directly
falsifiable prediction---that the word position of the first named entity should
track $t_{\mathrm{suf}}$ more tightly than question type does---which a lightweight
entity-tagging pass over the queries could test without any new retrieval. If that
correlation held, first-entity position would also furnish a far cheaper deployment
signal than the full prefix-retrieval sweep. We flag both as the natural next
measurement.

\subsection{RQ2: What fraction of queries admit latency hiding?}
\begin{figure}[t]
\centering
\includegraphics[width=0.78\linewidth]{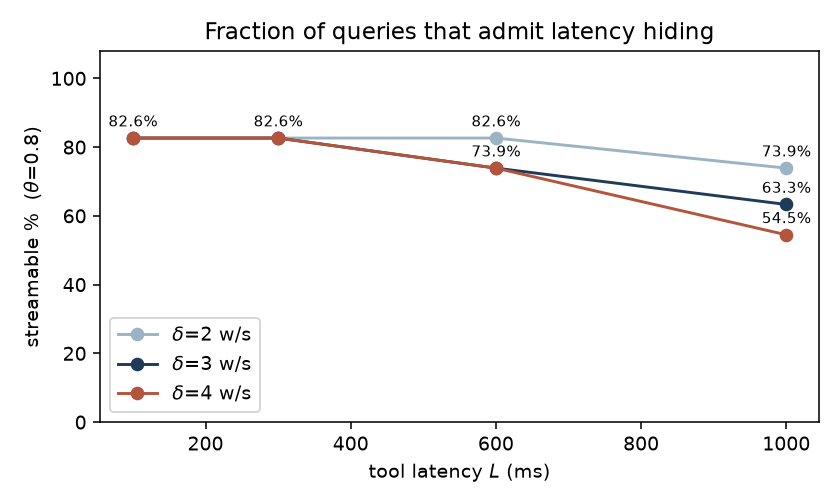}
\caption{Streamable fraction vs.\ tool latency $L$ at $\theta{=}0.8$, one line per
cadence $\delta$, with exact values annotated. The curves separate only once $L$ is
large enough for residual input time to bind; there, slower cadence dominates (H2).}
\label{fig:rq2}
\end{figure}

Using $t^\star=t_{\mathrm{suf}}$ where available and falling back to
$t_{\mathrm{sc}}$ otherwise, we compute the streamable fraction over the full
$(L,\delta,\theta)$ grid (Fig.~\ref{fig:rq2}). At the central
operating point ($L{=}600$\,ms, $\delta{=}3$\,w/s, $\theta{=}0.8$), \streamPct{} of
queries admit hiding at least 80\% of tool latency.

This blended figure mixes two populations and two notions of stabilization, and we
report it decomposed. On the \nRetrPct{} of queries with retrieved gold---where
$t^\star=t_{\mathrm{suf}}$ reflects evidence actually in hand---\streamSufPct{} are
streamable; on the remaining \nFallbackPct{}, where we fall back to the grounding-free
$t_{\mathrm{sc}}$ (top-1 stopped moving), \streamScPct{} are. The
sufficiency-only number, \streamSufPct{}, is the one that matches our framing and is
the honest headline for ``evidence is retrievable in time''; the \streamPct{} blend
is lower because it is dominated by the $t_{\mathrm{sc}}$ majority, for which
``streamable'' means only that the lexical top-1 has settled, not that gold is in
the top-$k$. Readers who care about evidence sufficiency should weight the
\streamSufPct{} figure; those who treat top-1 stability as the deployment signal get
the blend.

The blend also counts only upside. ``Streamable'' asks whether enough tool latency
\emph{can} be hidden; it is silent on whether the speculative query was
\emph{right}, and $H$'s $\max(0,\cdot)$ floor cannot represent the negative saving a
mis-fire incurs (\S5.4). One might expect this blind spot to bite hardest in the
$t_{\mathrm{sc}}$-fallback majority, which stabilizes \emph{late}
(mean $\phi{=}\phiScReplay{}$ vs.\ $\approx0.29$ on the retrieved-gold replay
sample), giving a fixed-interval trigger more room to fire before intent settles. We
tested this by replaying \rqThreeN{} fallback questions (the harness previously
replayed only retrieved-gold) through the same pipeline at $L{=}600$\,ms. The
expectation does \emph{not} hold: the net-negative-saving rate is \negRateSc{}
($1/\rqThreeN{}$), identical to the \negRateSuf{} on the favorable retrieved-gold
population, and mean savings are likewise comparable. The reason is the same constant
that dominates RQ3 (\S5.4): the query-generation overlap is hidden regardless of
whether the speculative \emph{retrieval} was right, so late stabilization lowers $H$
without proportionally raising the mis-fire rate at this operating point. Mis-fires
are thus real but rare ($\approx\negRateSc{}$) across both populations here; the
streamable fraction should still be read as an upper envelope on benefit paired with
this small measured downside, and the rate may grow at larger $L$ or with a more
aggressive trigger---a sweep we leave to future work.

Two further patterns stand out. First,
for small tool latency ($L\le300$\,ms) the constraint is essentially never binding:
$82.6\%$ of queries are streamable regardless of cadence---the residual input time
trivially covers a short tool call. Second, and confirming \textbf{H2}, when tool
latency is large ($L{=}1000$\,ms) the binding constraint becomes residual input
time, so a \emph{slower} cadence \emph{raises} the streamable fraction: from
$54.5\%$ at $\delta{=}4$\,w/s up to $73.9\%$ at $\delta{=}2$\,w/s. Faster typists
and faster talkers leave less room to hide a slow tool.

\subsection{RQ3: Does the bound match a working pipeline?}
\begin{figure}[t]
\centering
\includegraphics[width=0.62\linewidth]{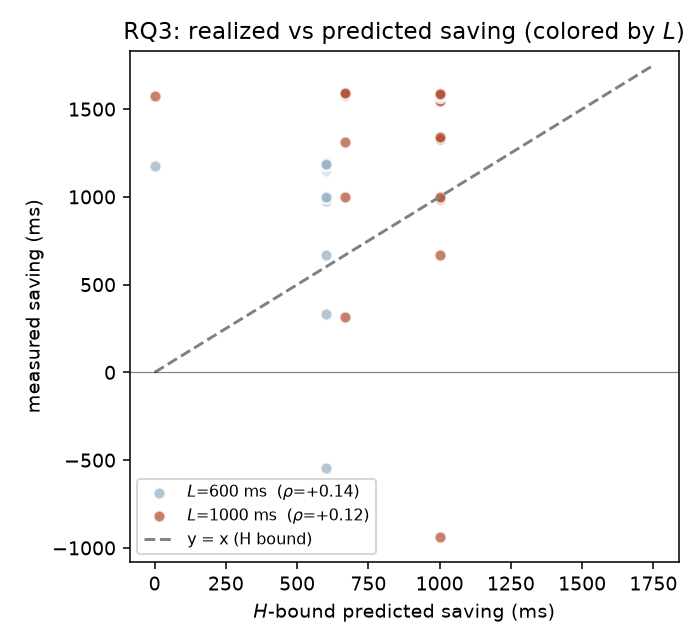}
\caption{RQ3: per-question measured saving vs.\ the $H$ bound for the same
\rqThreeN{} questions, colored by tool latency $L$. Most points lie above $y{=}x$
because the pipeline also hides query-generation latency, which $H$ does not model.
At $L{=}600$\,ms (light) the bound saturates---$\nSatLsix{}/\rqThreeN{}$ points share
$H{=}600$\,ms; at $L{=}1000$\,ms (dark) $H$ spreads across $\{0,667,1000\}$\,ms, yet
measured savings still do not track it (Spearman $\rho{=}\rhoLk{}$). The lone point
below zero is a late-stabilizing query whose mis-fire the floored bound cannot
express---and whose $H$ is nonetheless maximal.}
\label{fig:rq3}
\end{figure}

We replayed the same \rqThreeN{} retrieved-gold questions through the asynchronous
Streaming RAG harness at $\delta{=}3$\,w/s and \emph{two} tool latencies,
$L\in\{600,1000\}$\,ms (Fig.~\ref{fig:rq3}). In the aggregate the bound is
\emph{conservative} at both: measured perceived-latency savings (baseline minus
streaming) average \measSaved{}\,ms vs.\ the \hPred{}\,ms $H$ bound at $L{=}600$, and
\measSavedLk{}\,ms vs.\ \hPredLk{}\,ms at $L{=}1000$. The overshoot is expected and
informative: $H$ models only the tool latency hidden behind input, whereas the
calibrated pipeline additionally overlaps the LLM's query-generation step
($\approx\qgenMs{}$\,ms~\cite{arora2025streamrag}) with the user's input. Indeed the
overshoot is almost exactly that step: at $L{=}600$ the bound mean
($\hPred{}$\,ms) plus the query-generation constant ($\approx\qgenMs{}$\,ms) recovers
the measured mean ($\approx\measSaved{}$\,ms) to within noise---a useful sanity
check, but also a caution that this paragraph mostly re-derives $H$ plus a constant
rather than independently testing $H$ (we address that next). So a query
that $H$ deems streamable will, in practice, save \emph{at least} as much as
predicted---the property that makes $H$ safe to plan with.

We are deliberately careful about the stronger claim the bound does \emph{not}
support: per-query prediction. At $L{=}600$ the bound saturates---\nSatLsix{} of the
\rqThreeN{} queries share $H{=}600$\,ms (the input residual exceeds $L$)---leaving no
horizontal spread to test ranking, and the rank correlation between $H$ and the
measured saving is negligible (Spearman $\rho{=}\rhoLsix{}$). We therefore re-ran the
\emph{identical} questions at $L{=}1000$ specifically to break the saturation: $H$
now takes three distinct values ($0$, $667$, $1000$\,ms) with \nTailLk{} of \rqThreeN{}
queries below the cap. The spread appears, but the correlation does not:
$\rho{=}\rhoLk{}$ over all \rqThreeN{}, and exactly $0.00$ within the unsaturated tail.
Two structural reasons explain why raising $L$ cannot rescue per-query prediction.
First, the dominant component of realized saving---the constant
$\approx\qgenMs{}$\,ms query-generation overlap---is independent of $H$, so it adds a
near-uniform offset that swamps the cross-query signal $H$ encodes. Second, the
failure mode is \emph{negative}, and $H$'s $\max(0,\cdot)$ floor cannot represent it.
RQ3 thus confirms $H$ as a conservative aggregate bound but \emph{not} a per-query
ranker; a predictive model would need to add the query-generation overlap to the
hideable budget and replace the floor with a mis-fire penalty.\footnote{%
Formalizing the first correction without re-running anything: a tighter aggregate
bound $H' = \min(L,\,\mathrm{residual}) + \min(t_{\mathrm{qg}},\,\mathrm{residual})$
folds the query-generation step ($t_{\mathrm{qg}}\approx\qgenMs{}$\,ms), which the
pipeline overlaps with input regardless of retrieval correctness, into the hideable
budget. At $L{=}600$ this predicts $\approx\hPred{}+\qgenMs{}\approx1180$\,ms against
the measured \measSaved{}\,ms---recovering the overshoot to within noise. $H'$ is
still aggregate, not per-query: it adds a near-constant offset and so cannot rescue
the rank correlation, and it inherits the $\max(0,\cdot)$ floor that the negative
tail violates.}

The negative tail makes the second point concrete and reconnects volatility $V$ to
outcomes. The single query that \emph{loses} time---a 17-word comparison
($t^\star{=}13$)---costs $-543$\,ms at $L{=}600$ and $-940$\,ms at $L{=}1000$, yet its
$H$ is \emph{maximal} ($1000$\,ms): with $n-t^\star{=}4$ words of nominal residual the
bound rates it fully streamable while the fixed-interval trigger, firing before intent
settles at word~13, mis-retrieves and pays a re-fire penalty. Notably this query has
$V{=}0$, so volatility did \emph{not} flag it---the risk here is \emph{late}
stabilization under an early-firing trigger, not post-stabilization churn. This is a
single-query observation ($n{=}1$ net-negative across the replayed populations), so we
offer the reading as a mechanism-level intuition rather than a validated design rule:
it \emph{suggests} that an early-commit safeguard should gate on $\phi$ (how late
intent settles relative to the trigger cadence) at least as much as on $V$. Whether
$\phi$ in fact out-predicts $V$ for mis-fires requires a larger negative sample---a
trigger-aggressiveness and large-$L$ sweep that deliberately induces mis-fires---which
we leave to future work and do not claim to have established here.

\subsection{Robustness}\label{sec:robust}
\paragraph{Grounding (bounding the selection bias).} Our sharpest claims live on the
\nRetrPct{} retrieved-gold slice, whose selection plausibly biases stabilization
early (\S5.1). To bound that bias we re-derive $d^\star$ with a deliberately
\emph{relaxed} matcher---bag-of-content-token containment rather than contiguous
substring (\S\ref{sec:method})---which trades precision for recall and so expands the
groundable population. Groundability rises from \nGroundPct{} to \groundFuzzyPct{} and the
retrieved-gold rate from \nRetrPct{} to \retrFuzzyPct{} (a sixth more questions). If
the early-stabilization result were largely a selection artifact, admitting these
harder-to-ground questions should pull $\phi_{\mathrm{suf}}$ up sharply. It does not:
mean $\phi_{\mathrm{suf}}$ moves only from \phiSufMean{} to \phiSufFuzzy{}
(median $0.143\to0.154$). The true value is thus bracketed in $[\phiSufMean{},
\phiSufFuzzy{}]$---both early---and the by-type ordering is preserved, now on firmer
footing: \texttt{set}, the latest type, grows from $n{=}2$ to $n{=}\nSetFuzzy{}$ and
remains latest. We read the early-sufficiency finding as signal that survives a
$1.5\times$ population expansion, not as an artifact of the strict matcher. (Fuzzy
grounding is lower-precision, so it if anything biases $\phi_{\mathrm{suf}}$ \emph{down};
that the estimate barely moves is the reassuring direction.)

\paragraph{Retriever (top-$k$).}
Varying the sufficiency top-$k$ shifts absolute values but preserves the
qualitative picture (Table~\ref{tab:topk}). Larger $k$ surfaces the gold passage
for more questions (retrieved-gold rate $15.6\%\to25.0\%$) and stabilizes slightly
earlier (mean $\phi_{\mathrm{suf}}$ $0.327\to0.248$), as expected since a more
permissive top-$k$ is easier to satisfy. The self-consistency curve is unchanged by
$k$ (it depends only on the top-1), and the central streamable fraction is stable
near $73\text{--}74\%$. The early-sufficiency / late-self-consistency gap holds at
every $k$.

\begin{table}[t]
\centering
\caption{Top-$k$ robustness. $\phi_{\mathrm{sc}}$ (mean 0.67 / median 0.73) is
independent of $k$ and omitted.}
\label{tab:topk}
\begin{tabular}{lrrr}
\toprule
$k$ & retrieved-gold & mean $\phi_{\mathrm{suf}}$ & median $\phi_{\mathrm{suf}}$ \\
\midrule
1 & 15.6\% & 0.327 & 0.200 \\
3 & 21.3\% & 0.264 & 0.143 \\
5 & 25.0\% & 0.248 & 0.125 \\
\bottomrule
\end{tabular}
\end{table}

\paragraph{Retriever (dense).} Our sharpest concern is that BM25's reliance on
early lexical anchors---not a property of the queries---drives the low
$\phi_{\mathrm{suf}}$ we report. We test this directly by re-running the full sweep
with a dense bi-encoder (\texttt{all-MiniLM-L6-v2}, cosine over the same passages)
in place of BM25. The early-stabilization finding is not a lexical artifact: under
dense retrieval sufficiency stabilizes if anything \emph{later}, not earlier (mean
$\phi_{\mathrm{suf}}\,\phiSufMean{}\to\phiSufDense{}$, median $\phiSufMed{}\to\phiSufDenseMed{}$;
$\phi_{\mathrm{sc}}\,\phiScMean{}\to\phiScDense{}$), while the dense retriever surfaces gold
for \emph{more} questions (retrieved-gold $\nRetrPct{}\to\retrDensePct{}$). A semantic
retriever that does not key on early keywords thus still finds the answer-bearing
passage well before the query ends---median $\phi_{\mathrm{suf}}=\phiSufDenseMed{}$,
within the first quarter of the utterance. The by-type ordering is preserved
(Spearman $\rho{=}\rhoDenseType{}$ between the BM25 and dense $\phi_{\mathrm{suf}}$
rankings): \emph{aggregation} and \emph{comparison} remain among the earliest under
both, and the disagreement is confined to the small-$n$ / grounding-ambiguous cells
(\texttt{false\_premise}, \texttt{set}, \texttt{post-processing}). We read the dense
arm as converting our principal open confound into a positive robustness result:
the early-sufficiency signal is retriever-general, and BM25, if anything, is the
\emph{more} optimistic backbone.

% ===================================================================== %

\section{Conclusion}
Tool-intent stabilization is a query-intrinsic, model-agnostic quantity that
bounds when streaming tool use can help and by how much. On CRAG, sufficiency
stabilizes early on the groundable, retrievable subset (mean
$\phi_{\mathrm{suf}}$ bracketed to $[\phiSufMean{},\,\phiSufFuzzy{}]$ by exact and
relaxed grounding); question type gives a significant but coarse
early/late split (Kruskal--Wallis $p=\kwAllP{}$, $\varepsilon^2{=}\epsSq{}$); and $H$
is a conservative aggregate floor that realized savings meet or exceed, yielding a
\streamPct{} streamable fraction over the full benchmark.
Two limitations temper these claims, and we probed both rather than merely noting
them. First, the sufficiency results are conditional on a favorable \nRetrPct{}
slice; the relaxed-grounding arm brackets the population effect to under $0.02$ in
$\phi_{\mathrm{suf}}$ (\S\ref{sec:robust}), bounding the finding as signal rather
than artifact---though a fuller recovery (dense or learned grounding) remains open.
Second, our system validation establishes conservativeness in aggregate but
\emph{not} per-query prediction: at $L{=}1000$\,ms, realized savings do not track
$H$ (Spearman $\rho{=}\rhoLk{}$), because a constant query-generation overlap
dominates the saving and trigger mis-fires produce negative outcomes the floored
bound cannot represent (\S5.4). The downside is real but rare
($\approx\negRateSc{}$ net-negative at $L{=}600$). The natural next steps are to
close the prediction gap---adding the query-generation overlap and a mis-fire
penalty---and to regress $\phi$ on query features. Even so, the present results
already give system builders a principled, training-free basis for trigger design,
and a build/skip decision for a learned trigger.

\section*{Limitations}
Four assumptions bound how far the present numbers generalize; we take them in turn,
from the input model through to the retriever and benchmark.
\paragraph{Input timing.} A uniform word stream is a proxy for speech; real ASR
pacing is variable and partial hypotheses get revised. We bound this here by using
clean text and treat speech timing as planned follow-up.
\paragraph{Stabilization measure.} Top-1 equality is a coarse proxy for ``intent
settled.'' We mitigate by also reporting sufficiency against the gold document and
by varying top-$k$.
\paragraph{Grounding.} $d^\star$ is derived by string matching, so the groundable
population excludes non-verbatim answers; reported $\phi_{\mathrm{suf}}$ statistics
are conditional on groundability, and we report $t_{\mathrm{sc}}$ as a
grounding-free check.
\paragraph{Fallback population.} On the \nFallbackPct{} of queries where BM25 never
surfaces the gold passage from any prefix, stabilization is measured only by
$t_{\mathrm{sc}}$ (when the top-1 stops moving). If the top-1 in this population is
the wrong document, early settling is \emph{stable mis-retrieval}, not evidence that
the speculative query was right: $t_{\mathrm{sc}}$ certifies that intent has
\emph{stopped changing}, not that it is \emph{correct}. This is the central gap
between our measurement and end-to-end streaming-RAG quality, and it is why we keep
the sufficiency and self-consistency populations separate throughout (\S5.3) rather
than reading the blended streamable fraction as an evidence-quality claim. Our RQ3
replay finds the realized-saving and net-negative rates comparable across the two
populations at $L{=}600$ (\S5.3), but that probes \emph{perceived latency}, not
retrieval correctness, on the fallback set.

\paragraph{Retriever and benchmark.} BM25 is lexical and phrasing-sensitive, and is
known to retrieve on early topical anchors. Because $t_{\mathrm{suf}}$ asks only when
the gold page enters the top-$k$, that early-anchor behavior could itself drive the
low $\phi_{\mathrm{suf}}$ we report. We addressed this head-on with the
dense-retriever arm (\S\ref{sec:robust}): a bi-encoder that weights terms
semantically rather than lexically surfaces gold for \emph{more} questions yet
stabilizes if anything \emph{later}, and the by-type ordering is preserved
($\rho{=}\rhoDenseType{}$)---so the early-stabilization claim is retriever-general,
not a BM25 artifact. What remains open is benchmark generality: a single
English QA set leaves cross-dataset and cross-lingual replication---where verb-final
word order may shift $t_{\mathrm{suf}}$ under lexical but not dense retrieval---as
the natural next step, not optional polish.

% \bibliographystyle is set by acl.sty; redeclaring it triggers BibTeX's
% "Illegal, another \bibstyle command" error.
\bibliography{refs}

\end{document}